\documentclass[letterpaper, 10 pt, journal, twoside]{IEEEtran}
\usepackage{amsmath,amsfonts}
\usepackage{algorithmic}
\usepackage{algorithm}
\usepackage{array}
\usepackage[caption=false,font=normalsize,labelfont=sf,textfont=sf]{subfig}
\usepackage{textcomp}
\usepackage{stfloats}
\usepackage{url}
\usepackage{verbatim}
\usepackage{graphicx}
\usepackage{cite}
\usepackage{hyperref}
\DeclareMathOperator*{\argmax}{arg\,max}

\newcommand{\cT}{\mathcal{T}}
\begin{document}

\title{Learning to Build: Autonomous Robotic Assembly of Stable Structures Without Predefined Plans}

\author{Jingwen Wang$^{*}$$^{1}$, Johannes Kirschner$^{*}$$^{2}$, Paul Rolland$^{*}$$^{2}$, Luis Salamanca$^{2}$, and Stefana Parascho$^{1}$% <-this % stops a space
\thanks{*Equal Contribution}% <-this % stops a space
\thanks{Manuscript received: April 24, 2026; Accepted Oct 19, 2026.}
\thanks{This paper was recommended for publication by Associate Editor C.C.Smith and Editor Ashis Banerjee upon evaluation of the Associate Editor and Reviewers’ comments. The development of the present work has been supported by research project C22-06 Robotoddler - Autonomous Robotic Assemblies (Lab for Creative Computation, EPFL, and Swiss Data Science Center, ETH Zurich)}
\thanks{$^{1}$ Lab for Creative Computation, EPFL, Switzerland\\
        {\tt\footnotesize jingwen.wang@epfl.ch}
        }%
\thanks{$^{2}$ Swiss Data Science Center, Switzerland
        % {\tt\small b.d.researcher@ieee.org}
        }%
\thanks{Digital Object Identifier (DOI):
\href{https://doi.org/10.1109/LRA.2026.3747578}
{10.1109/LRA.2026.3747578}}
}

% \thanks{Manuscript received April 19, 2026; revised June 12, 2026.}}

\markboth{IEEE ROBOTICS AND AUTOMATION LETTERS, PREPRINT VERSION. October, 2026}%
{Wang \MakeLowercase{\textit{et al.}}: Learning to Build: Autonomous Robotic Assembly of Stable Structures Without Predefined Plans}

\IEEEpubid{%
  \begin{minipage}{\textwidth}
    \vspace*{\baselineskip}
    \footnotesize
    \raggedright
    \copyright\ 2026 IEEE. Personal use of this material is permitted.
    Permission from IEEE must be obtained for all other uses, in any
    current or future media, including reprinting/republishing this
    material for advertising or promotional purposes, creating new
    collective works, for resale or redistribution to servers or lists,
    or reuse of any copyrighted component of this work in other works.
  \end{minipage}%
}

\maketitle

\begin{abstract}
This paper presents a novel autonomous robotic assembly framework for constructing stable structures without relying on predefined architectural blueprints. Instead of following fixed plans, construction tasks are defined through targets and obstacles, allowing the system to adapt more flexibly to environmental uncertainty and variations during the building process. A reinforcement learning (RL) policy, trained using deep Q-learning with successor features, serves as the decision-making component. As a proof of concept, we evaluate the approach on a benchmark of 15 2D robotic assembly tasks of discrete block construction. Experiments using a real-world closed-loop robotic setup demonstrate the feasibility of the method and its ability to handle construction noise. The results suggest that our framework offers a promising direction for more adaptable and robust robotic construction in real-world environments.
\end{abstract}


%%%%%%%%%%%%%%%%%%%%%%%%%%%%%%%%%%%%%%%%%%%%%%%%%%%%%%%%%%%%%%%%%%%%%%%%%%%%%%%%
\section{Introduction}
\label{sec:intro}

\IEEEPARstart{I}{n} the past decade, robots have been increasingly adopted in construction~\cite{autodesk_2025} for their precision and efficiency, yet their widespread use remains limited. A major barrier is the reliance on rigid, highly detailed plans that are difficult to apply in the dynamic and uncertain conditions of construction sites~\cite{ parascho_construction_2023, liu_robotics_2024}. Unlike manufacturing, where robotic arms execute repetitive tasks in controlled environments using pre-tested motion plans, architectural fabrication faces unique and unpredictable on-site conditions—uneven terrain, variable material properties, and human-induced inconsistencies. As a result, plan-driven workflows that assume perfect alignment between design and reality often prove impractical, demanding tight tolerances and leaving little room for material or fabrication uncertainties or real-time adaptation~\cite{love_state_2022}. These challenges highlight the need for construction systems that can make decisions on the fly, adjusting their actions as conditions evolve instead of relying solely on precomputed plans.

Building on this motivation, we aim to develop a construction framework that enables robots to autonomously perform construction tasks without relying on predefined plans. 
We focus specifically on the robotic assembly of discrete rigid block structures with dry joints. This restriction allows us to target architectural assemblies that can be built through a relatively straightforward and resource-efficient construction process, as commonly explored in research on unreinforced or dry-stacked masonry systems~\cite{block_beyond_2017, dorfler_multi-robotic_2023}. %However, because dry joints do not provide inherent reinforcement, structural stability must be maintained throughout the construction process, which makes adaptive decision-making particularly important.

% Specifically, this work investigates: 
% \begin{itemize}
%     \item How to formalize robotic assembly tasks in a way that frees the process from rigid, predefined construction plans.
%     \item How to design a flexible and adaptable framework capable of accomplishing multiple construction tasks.
%     \item How to validate the proposed framework in physical construction scenarios with noise.
% \end{itemize}

    % In this paper, we propose a flexible robotic construction system that builds structures from discrete blocks based on abstract task definitions, specified through targets and obstacles, without utilizing predefined plans, as shown in Figure \ref{fig:problemdefinition} (left). A reinforcement learning (RL) model serves as the decision-making core, enabling the robot to adapt its actions based on construction progress and real-time site conditions. As a proof of concept, we demonstrate our approach on a set of 2D dry-stacked vertical and spanning structures, and validate it both in simulation and on a real-world closed-loop robotic assembly process.

In summary, our work makes the following contributions:

\begin{itemize}
    % \item We propose a novel robotic construction framework that enables the autonomous assembly of 2D stable structures without reliance on predefined architectural plans, as shown in Fig~\ref{fig:problemdefinition} (left).

    % \item We formalize construction tasks in terms of geometric goals, defined by targets and obstacles, rather than fixed geometric forms, allowing structural layouts to emerge during the building process.

    % \item We develop a goal-conditioned reinforcement learning (RL) approach that leverages successor features with deep Q-learning to solve multiple construction tasks requiring structural stability during construction. 

    % %\item Show that the learned RL policy can dynamically adapt construction decisions in response to environmental changes, ensuring robust performance under real-world variability.
    
    % %\item Demonstrate the transferability of the trained policies across different construction scenarios, highlighting generalization capabilities.

    % \item We validate the framework in simulation and through real-world closed-loop robotic demonstrations.

    \item We propose a blueprint-free robotic construction framework in which tasks are specified by targets and obstacles, allowing structural layouts to emerge during construction.
    \item We develop a goal-conditioned RL approach based on deep Q-learning with successor features and validate it on diverse 2D construction tasks in simulation and closed-loop robotic experiments.

\end{itemize} 

\section{Related Works}
\label{sec:relatedwork}

\subsection{Classic plan-based robotic assembly of stable structures}

Existing research on robotic assembly of dry-stacked structures~\cite{parascho_robotic_2020, nagele_legobot_2020, dorfler_multi-robotic_2023} typically depends on carefully analyzed construction plans, detailing geometry, assembly sequence, and robotic execution. This linear design-to-construction pipeline often results in rigid workflows with limited adaptability.

Three practical challenges further illustrate these limitations. First, identifying a feasible and stable assembly sequence for a certain geometry, particularly for spanning structures with strict stability constraints, is an NP-hard problem~\cite{deuss_assembling_2014, wang_learning_2025}. Second, robotic constraints must also be taken into account: even if a sequence satisfies structural stability requirements, it may still be unconstructible due to issues such as limited robot reachability or potential collisions with previously placed elements or the robot itself~\cite{nagele_legobot_2020, dorfler_multi-robotic_2023, wang_learning_2025}. These constraints often necessitate revisiting early design decisions. Third, construction environments are inherently dynamic. Construction tolerances tend to accumulate, especially in tall or complex assemblies \cite{dorfler_multi-robotic_2023}. Such variability further undermines the feasibility of rigid plans.

To address these challenges, we aim to develop a system that moves beyond static, plan-based construction and can dynamically adapt to on-site conditions.

\subsection{Learning-based assembly guided by blueprints}

\IEEEpubidadjcol
Recent work has demonstrated the use of learning-based methods to assemble structures from predefined blueprints. Ghasemipour et al.~\cite{ghasemipour_blocks_2022} train agents to assemble magnetic structures using structured graph representations and reinforcement learning, learning both sequencing and bimanual coordination. Kulshrestha and Qureshi~\cite{kulshrestha_structural_2023} use graph attention networks to plan multi-level rearrangements of up to 20 blocks, then delegating low-level actions to a robot controller for demonstration. Funk et al.~\cite{funk_learn2assemble_2021} propose a hierarchical system combining graph-based policies with Monte Carlo Tree Search to construct stable 3D structures consisting of cubes. 

These works capture structural dependencies such as layering~\cite{kulshrestha_structural_2023, funk_learn2assemble_2021} and construction coordination strategies~\cite{ghasemipour_blocks_2022, kulshrestha_structural_2023}, yet they all operate under fixed designs. In contrast, our work addresses the challenge of constructing without fixed designs, requiring the agent to determine how to achieve a goal under evolving and unstructured conditions.

\subsection{Learning-based assembly without blueprints}

Approaches without predefined plans often target specific structural forms.  Li et al.~\cite{li_learning_2022} learn to build bridges over gaps using attention-based networks and object-centric policies. Vallat et al.~\cite{vallat_reinforcement_2023} train agents to construct arches across variable spans. Both show that learning-based methods can flexibly generate bridge-like structures from task-specific goals.  Wibranek et al.~\cite{wibranek_reinforcement_2021} learn to assemble SL blocks along curves to construct linear structures sequentially. Bapst et al.~\cite{bapst_structured_2019} propose a graph-based approach that learns construction strategies through relational reasoning. One of their tasks is to connect the ground to targets while avoiding obstacles, allowing flexible structural layouts. However, their task design often produces similar umbrella-like solutions, and the use of adhesive connections leads to behaviors that do not generalize to dry-stacked construction.

Unlike prior approaches that focus on a single structural form, our framework is designed to handle a wider range of construction goals, supporting both vertical and spanning topologies. It also departs from the common assumption of rectangular or grid-aligned blocks~\cite{li_learning_2022, bapst_structured_2019, vallat_reinforcement_2023, wibranek_reinforcement_2021}. By accommodating polygonal geometries such as trapezoids, our system enables more diverse and flexible construction strategies.

\section{Problem Formulation}
\label{sec:problem}
\subsection{Task overview and design principles}
\label{subsec:task}
\begin{figure*}[t]
    \centering
    \includegraphics[width=6.9in]{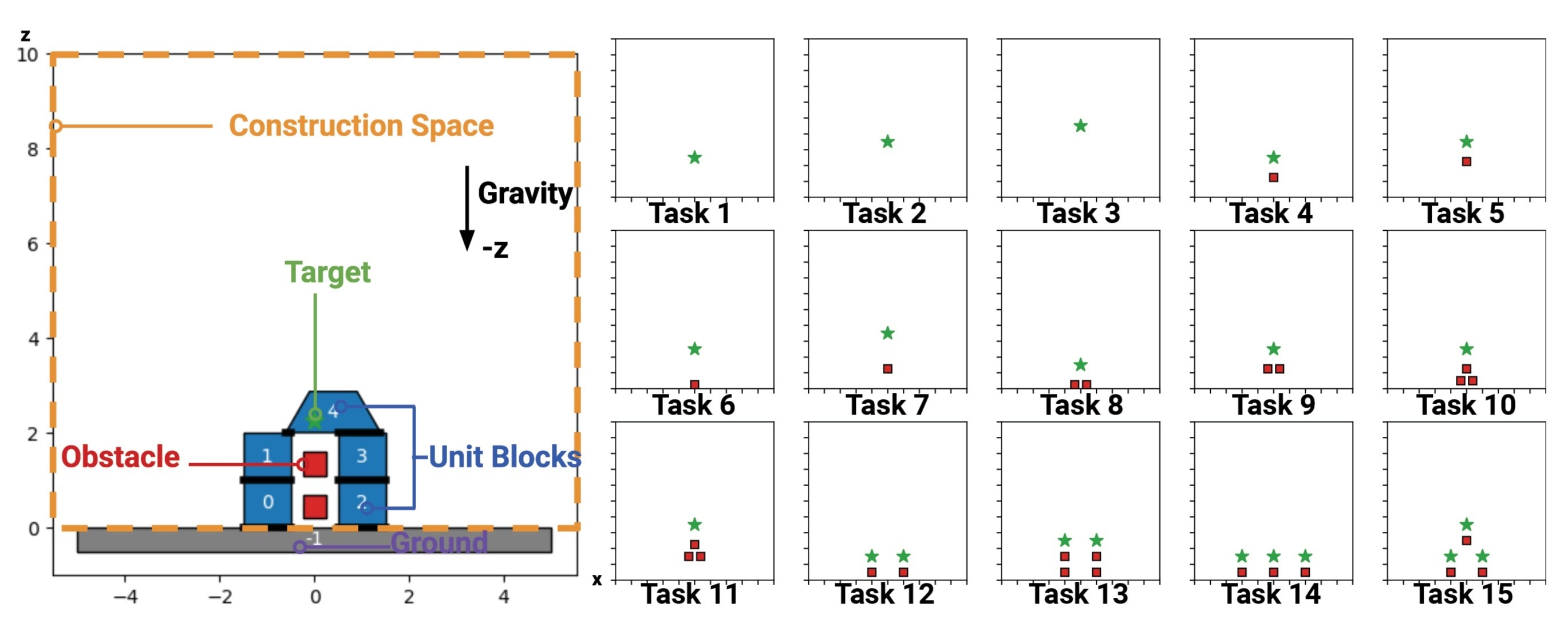}
    \caption{\textbf{Left:} Problem Definition: Example of a construction task, defined by the construction space, target locations, obstacle regions, and available unit blocks. \textbf{Right:} Tasks: A set of 15 tasks used to evaluate the reinforcement learning approach.}
    \label{fig:problemdefinition}
    \vspace{-1.5em}
    \end{figure*}

To support the construction of diverse architectural forms, we define a construction task in terms of four elements: the construction space, targets, obstacles, and blocks. As shown in Fig~\ref{fig:problemdefinition}, the goal of the task is for the robot to build a discrete structure by using a set of blocks that connect the ground to the targets while avoiding obstacles. This formulation offers several key advantages. First, it enables the construction of a wide range of structural forms simply by varying the configuration of targets and obstacles. For example, a column or a bridge can be represented through different task specifications. Second, it removes the need to provide a fixed blueprint for the robot to follow, allowing fast adaptation to unexpected changes or inaccuracies. 

\subsection{Simulated environment definition}
\label{subsec:env}
To simulate the block assembly process in Section~\ref{subsec:task}, we design a 2D construction environment that captures essential physical and geometric constraints relevant to architectural assembly tasks, shown in Fig~\ref{fig:problemdefinition}(left). In this environment, the most relevant elements are:

\textbf{Blocks.}
We use two basic types of construction units: square and trapezoidal blocks. These shapes are commonly found in architectural applications and provide diversity for creating varied structures. 
Square blocks have side length 1, while trapezoidal blocks have edge lengths $1,1,1,$ and $2$. Blocks are placed only through edge-to-edge contact, with their orientation determined by the selected contacting edges. The inventory is unlimited during simulation.

\textbf{Construction Space.}  
The construction space is defined based on the dimensions of the blocks. It spans the range $x \in [-5, 5]$ and $z \in [0, 10]$. A horizontal floor is placed at $z = 0$, serving as the foundation upon which all structures are built.

\textbf{Targets and Obstacles.} 
Targets are defined as 2D points that the structure must reach. Obstacles are defined as 2D square regions that the structure must avoid touching. These elements encode high-level construction objectives and constraints within each task.

Furthermore, we can define the most important concepts related to the construction process:

\textbf{Construction Step.}  
At each construction step, a new block is placed within the environment. A placement action is considered valid if the block does not overlap with any other block already placed, the ground floor and obstacles, the block remains within the bounds of the construction space, and the resulting structure remains stable. Structural stability is evaluated using the Rigid-Block Equilibrium (RBE) method~\cite{Whiting_Procedural_2009}. The construction episode, defined as the complete sequence of these construction steps, terminates successfully when all targets have been reached; otherwise, it terminates unsuccessfully if the maximum number of blocks has been placed.

\textbf{Construction Task.}  
A construction task is defined by a set of target points and obstacle regions defined by squares. To promote the emergence of specific structural typologies—such as columns, arches, and bridges—we manually design a set of tasks, illustrated in Fig~\ref{fig:problemdefinition}(right).

%Block shapes 

%Different construction tasks (shown in Figure \ref{fig:task})

    % \begin{figure}
    % \centering
    % \includegraphics[width=5in]{Figures/tasks.pdf}
    % \caption{A set of 15 tasks used to evaluate the reinforcement learning approach.}
    % \label{fig:task}
    % \end{figure}

\subsection{Challenges of solving the proposed problem}

%To the best of our knowledge, this construction problem has not been explored previously. Prior work generally assumes fixed designs or targets a single structural topology, whereas our goal is to solve a diverse set of tasks with one policy.

Our construction problem is closely related to prior blueprint-free construction tasks\cite{bapst_structured_2019}, but focuses on solving a diverse set of dry-stacked construction tasks with one policy.

The main challenge lies in the dynamic, state-dependent action space: each block placement changes the geometry, creating a new set of valid placements that grows combinatorially as the structure expands. The agent must reason over this evolving configuration to plan long-horizon assembly sequences. In addition, the reward varies across tasks due to different target and obstacle layouts. As a result, the learning framework must enable a policy that handles a continuously changing action space while also generalizing across tasks with distinct objectives.
\section{Methodology}
\label{sec:method}
% \subsection{Overall workflow}
To address the proposed construction problem, we develop a goal-conditioned reinforcement learning framework and validate it in a real-world closed-loop robotic system.

\newcommand{\tfinal}{t_{\text{final}}}   

\subsection{Goal-conditioned formulation}

We define the construction task as a goal-conditioned reinforcement learning problem, defined by a state space $\mathcal{S}$, action space $\mathcal{A}$, and task space $\mathcal{T}$~\cite{andrychowicz2017hindsight}. The goal-conditioned formulation enables a single construction policy to condition its decisions on different construction tasks and potentially generalize to unseen task configurations. A task $T \in \mathcal{T}$ specifies targets, obstacles, and available block shapes. The state $S_t = \{A_1, \dots, A_t\}$ represents the current assembly. At each step, $\mathcal{A}(S_t)$ denotes the set of feasible block placements. Candidate placements are generated by aligning an edge of a new block with an edge of an already placed block and sampling five relative offsets along the contact range. Candidates that collide with existing blocks, obstacles, or the ground are removed, followed by those classified as unstable by the RBE solver. The learned policy then selects an action $A_{t+1} \in \mathcal{A}(S_t)$ from this feasible action set, which is added to the assembly.
Given a task $T$, the goal is to construct an assembly that reaches all targets without intersecting any obstacles. An episode terminates when all targets are reached, no feasible actions remain, or the maximum number of steps is exceeded.

% Jingwen: I removed some details to save space

\subsection{RL formulation}
Designing an RL model for our construction tasks requires addressing three key aspects. 
First, we must encode the state space $\mathcal{S}$, the action space $\mathcal{A}$, and the task space $\mathcal{T}$ in a way that effectively captures the combinatorial assembly space, construction goals, and geometric properties of the construction. 
Second, the reward formulation should guide the assembly toward targets efficiently—using as few blocks as possible—while enabling multi-task learning. Third, the learning algorithm must handle a state-dependent action space, as the set of feasible placements $\mathcal{A}(S)$ changes with the evolving structure and cannot be captured by a fixed global action set.

\begin{figure}[t]
    \centering
    \includegraphics[width=3in]{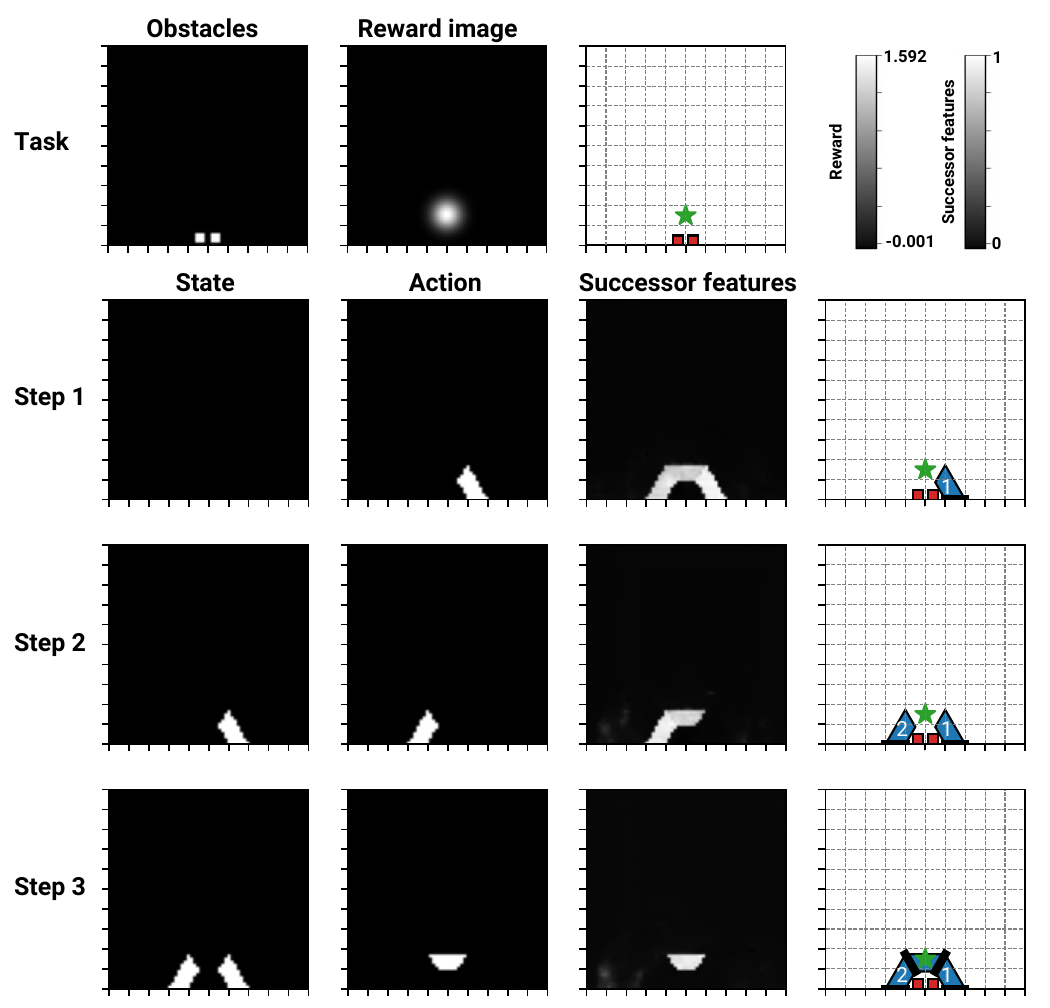}
    \caption{Example episode for Task 8. The successor-feature images visualize the discounted future block occupancy predicted for the selected action, providing a spatial representation of the policy's anticipated construction. Reward and successor-feature values are dimensionless.}
    \label{fig:episode_run}
    \vspace{-1.5em}
    \end{figure}

\textbf{Feature representation.}
To address the first point, we adopt a shared image-based representation for states, actions, and tasks. Each action is encoded as a binary image $\phi(A) \in \mathbb{R}^{d \times d}$ (Fig~\ref{fig:episode_run} Action Column), and the state is represented as an additive composition $\psi(S_t) = \sum_{i=1}^t \phi(A_i)$ (Fig~\ref{fig:episode_run} State Column), reflecting the incremental assembly process. The task is encoded as $\xi(T) \in \mathbb{R}^{2 \times d \times d}$ (Fig~\ref{fig:episode_run} Task Row), where the two channels correspond to obstacles and targets. In our experiments, we set $d = 64$, capturing sufficient geometric detail for our benchmark tasks (Fig \ref{fig:problemdefinition}).

Importantly, this formulation ensures that state, action, and task features all share a common image-based representation.

\textbf{Reward.}
Second, to guide the block placement, we introduce a task-dependent reward $r(A, T)$. A naïve formulation would assign a reward of $+1$ only when a block is placed at a target location, but such a sparse reward severely slows learning. To address this issue, we replace the single-point target signal with a smooth scalar field around the targets, encouraging the structure to grow toward them. In addition, we penalize the volume of each placed block to promote material efficiency. 
For each task, we construct a reward feature $\rho(T)\in \mathbb{R}^{64 \times 64}$ by convolving the target locations with two Gaussian kernels with standard deviations $\sigma_1=1$ and $\sigma_2=0.1$ and summing the resulting fields. A constant penalty of $5/N$, where $N$ is the number of image pixels, is subtracted from the reward field to penalize block usage. Then, the reward is given by
\begin{equation}
r(A,T) = \phi(A)^\top \rho(T),
\end{equation}
where $\phi(A)$ denotes the action feature. This formulation provides dense, shaped rewards and enables fast evaluation by decomposing the reward into task-dependent and action-dependent components directly in the shared image feature space.

\textbf{Learning.}
Third, we must choose an effective learning algorithm for the RL setup described above. Our objective is to learn a policy $\pi : \mathcal{S} \times \mathcal{T} \rightarrow \mathcal{A}$ maximizing the discounted return
\begin{equation}
V^\pi = \sum_{t=0}^{\tfinal - 1} \gamma^{t} r(A_{t+1}, T),
\end{equation}
where $\gamma \in [0,1]$ is the discount factor and $\tfinal$ is the number of steps until the episode ends.
We exploit the linear structure of the reward $r(A,T) = \phi(A)^\top \rho(T)$ and introduce the \textit{successor features} $\Psi^{\pi}(S,A,T)$. Rather than directly mapping $(S,A,T)$ to a scalar action value, this parameterization separates the anticipated spatial consequence of an action from its task-dependent evaluation. For a trajectory starting from $(S,A,T)$, we define the successor features as
\begin{equation}
\Psi^{\pi}(S,A,T) = \phi(A) + \sum_{t > |S|} \gamma^{t - |S|} \phi(A_{t+1}),
\end{equation}
so that the state-action value function decomposes as
\begin{equation}
Q^\pi(S,A,T) = \Psi^{\pi}(S,A,T)^\top \rho(T).
\end{equation}
In our image-based formulation, $\Psi^\pi(S,A,T)$ represents the discounted future block occupancy expected after selecting action $A$, while $\rho(T)$ encodes the task objective through the target-dependent reward field. Thus, the network first predicts the spatial consequence of a placement and only then evaluates this prediction against the task reward. In contrast to directly predicting a scalar $Q(S,A,T)$, this factorization explicitly exposes the anticipated future construction and provides a spatially interpretable representation of the policy's long-term behavior.

Optimal successor features are known to satisfy a Bellman equation analogous to that of value and Q-functions~\cite{barreto2017successor}, which allows us to apply standard RL methods in this feature space. As a result, instead of directly approximating $Q^\pi$, our RL algorithm learns to predict the successor features of the optimal policy.

We approximate $\Psi_\theta(S,A,T)$ using a U-Net~\cite{ronneberger2015u} and train it with a variant of Deep Q-Learning~\cite{mnih2015human}. The network takes a $4\times64\times64$ input and uses three encoder stages with 16, 32, and 64 channels, respectively. Each stage contains two $3\times3$ convolutions with ReLU activations; the first two stages are followed by $2\times2$ max pooling. The decoder uses the corresponding transposed operations following a standard U-Net architecture. No additional exploration mechanism, such as $\epsilon$-greedy exploration, is used; actions are selected greedily according to the current value estimate.
The complete procedure is summarized in Algorithm 1. 
After training, an episode run resembles what is illustrated in Fig~\ref{fig:episode_run}. Column 3 corresponds to the successor feature $\Psi_\theta(S,A^*,T)$ which is used to select the action $A^*$ in this step, i.e., $A^* = \argmax_{A} \Psi_\theta(S,A,T)^\top \rho(T)$. We can see that Column 3, the trained successor features, also show the future assembly to be constructed. 

\begin{figure}[t]
\vspace{0.4em}
\centering
\footnotesize
\fbox{%
\begin{minipage}{0.46\textwidth}
\textbf{Algorithm 1.} Deep Q-Learning with Successor Features

\vspace{0.3em}
\textbf{Input:} task set $\cT=\{T_1,\dots,T_k\}$, batch size $B$, episodes $M$, 
optimization epochs $n_{\mathrm{optim}}$, policy iterations $n_{\mathrm{policy}}$

\textbf{Initialize:} replay buffer $\mathcal{D}$; network $\Psi_\theta$; target network $\Psi_{\bar{\theta}}$ with $\bar{\theta}\leftarrow\theta$

\vspace{0.3em}
\textbf{for} episode $=1,\dots,M$ \textbf{do} \textit{  // collect transitions}

\hspace*{1em}\textbf{for each} $T\in\cT$ \textbf{do}

\hspace*{2em}$S\leftarrow S_0$

\hspace*{2em}\textbf{for} $t=1,\dots,\text{max steps}$ or until termination \textbf{do}

\hspace*{3em}$A\leftarrow \arg\max_{A'} \Psi_\theta(S,A',T)^\top \rho(T)$

\hspace*{3em}execute $A$, observe $S'$

\hspace*{3em}store $(S,A,S',T)$ in $\mathcal{D}$

\hspace*{3em}$S\leftarrow S'$

\hspace*{2em}\textbf{end for}

\hspace*{1em}\textbf{end for}

\hspace*{1em}\textbf{for} $1,\dots,n_{\mathrm{policy}}$ \textbf{do} \textit{  // update policy}

\hspace*{2em}\textbf{for} $1,\dots,n_{\mathrm{optim}}$ \textbf{do}

\hspace*{3em}sample batch $\{(S_j,A_j,S_j',T_j)\}_{j=1}^B$ from $\mathcal{D}$

\hspace*{3em}\textbf{for each} sampled transition \textbf{do}

\hspace*{4em}$A_j' \leftarrow \arg\max_{A'} \Psi_{\bar{\theta}}(S_j',A',T_j)^\top \rho(T_j)$

\hspace*{4em}\textbf{if} $S_j'$ non-terminal:
\[
Y_j=\phi(A_j)+\gamma \Psi_\theta(S_j',A_j',T_j)
\]
\hspace*{4em}\textbf{else} $Y_j=\phi(A_j)$

\hspace*{3em}\textbf{end for}

\hspace*{3em}update $\theta$ with ADAM on
\[
L(\theta)=\frac{1}{B}\sum_{j=1}^B \|Y_j-\Psi_\theta(S_j,A_j,T_j)\|^2
\]

\hspace*{2em}\textbf{end for}

\hspace*{2em}$\bar{\theta}\leftarrow \theta$

\hspace*{1em}\textbf{end for}

\textbf{return} $\pi_\theta(S,T)=\arg\max_A \Psi_\theta(S,A,T)^\top \rho(T)$
\end{minipage}}
% \caption{Training procedure for the proposed DQN with successor features.}
\label{alg:1}
\vspace{-1.5em}
\end{figure}

    \begin{figure}[t]
    \vspace{1em}
    \centering
    \includegraphics[width=3in]{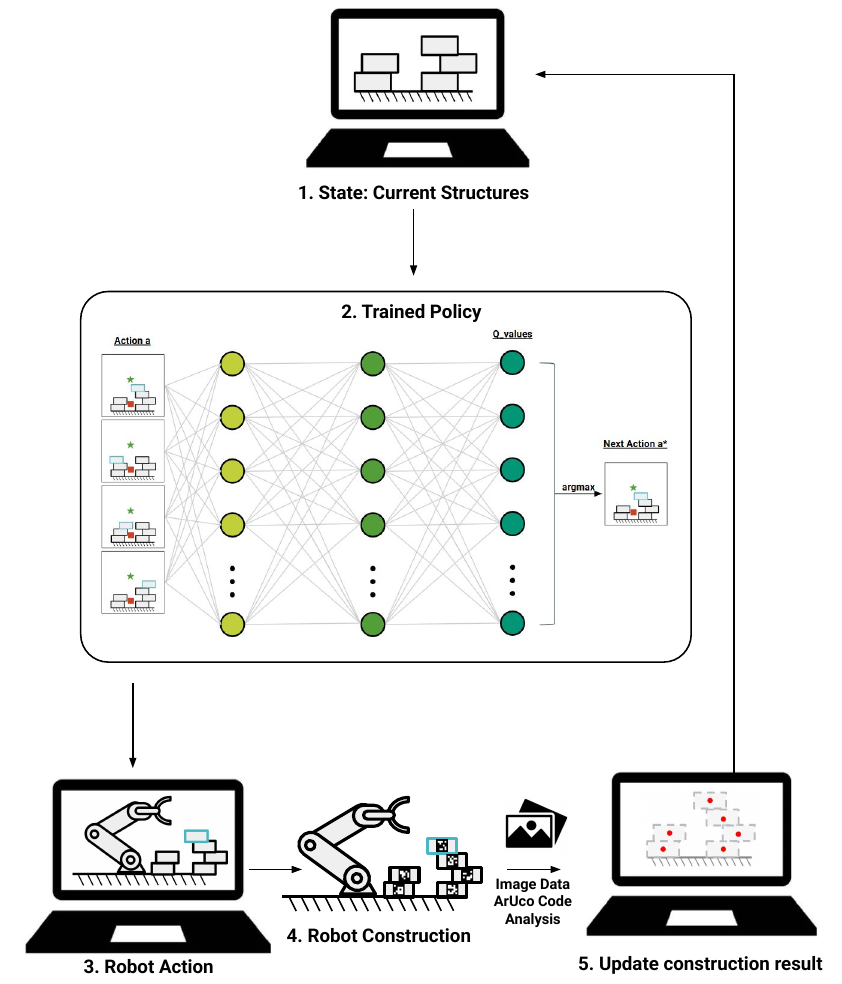}
    \caption{Closed-loop robotic assembly workflow. Given the current constructed structure (state $S$) and the task specification, the trained policy selects the next action $A$ for the robot to execute. After the robot places the block in the physical world, ArUco-based pose estimation is used to detect the updated block configuration. This information is then fed back into the simulation to update the state $S$ for the next decision step.}
    \label{fig:workflow}
    \vspace{-1.5em}
    \end{figure}

\subsection{Robotic closed-loop construction}

% We implement a closed-loop robotic assembly system to validate the trained reinforcement learning policy and its capacity to react to unplanned changes due to construction noise.

We deploy the learned policy in a closed-loop robotic system to evaluate robustness under real-world noise. The policy selects a desired block placement from the feasible action set, while the robotic system converts this placement into low-level execution. Closed-loop feedback is crucial, as physical construction introduces small placement errors and tolerances that accumulate over time. By updating the state through real-time perception, the system allows the policy to adapt its decisions to the actual construction state.

% We implement a closed-loop robotic assembly system to evaluate whether the trained RL policy remains effective under physical assembly noise. Closed-loop feedback is crucial, as physical construction introduces small placement errors and tolerances that accumulate over time. By updating the state through real-time perception, the system allows the policy to adapt its decisions to the actual construction state. This serves as a proof of concept, demonstrating that our framework can accommodate fabrication uncertainties and supports real-time adaptation.

\begin{figure}[t]
    \vspace{0.3em}
    \centering
    \includegraphics[width=3in]{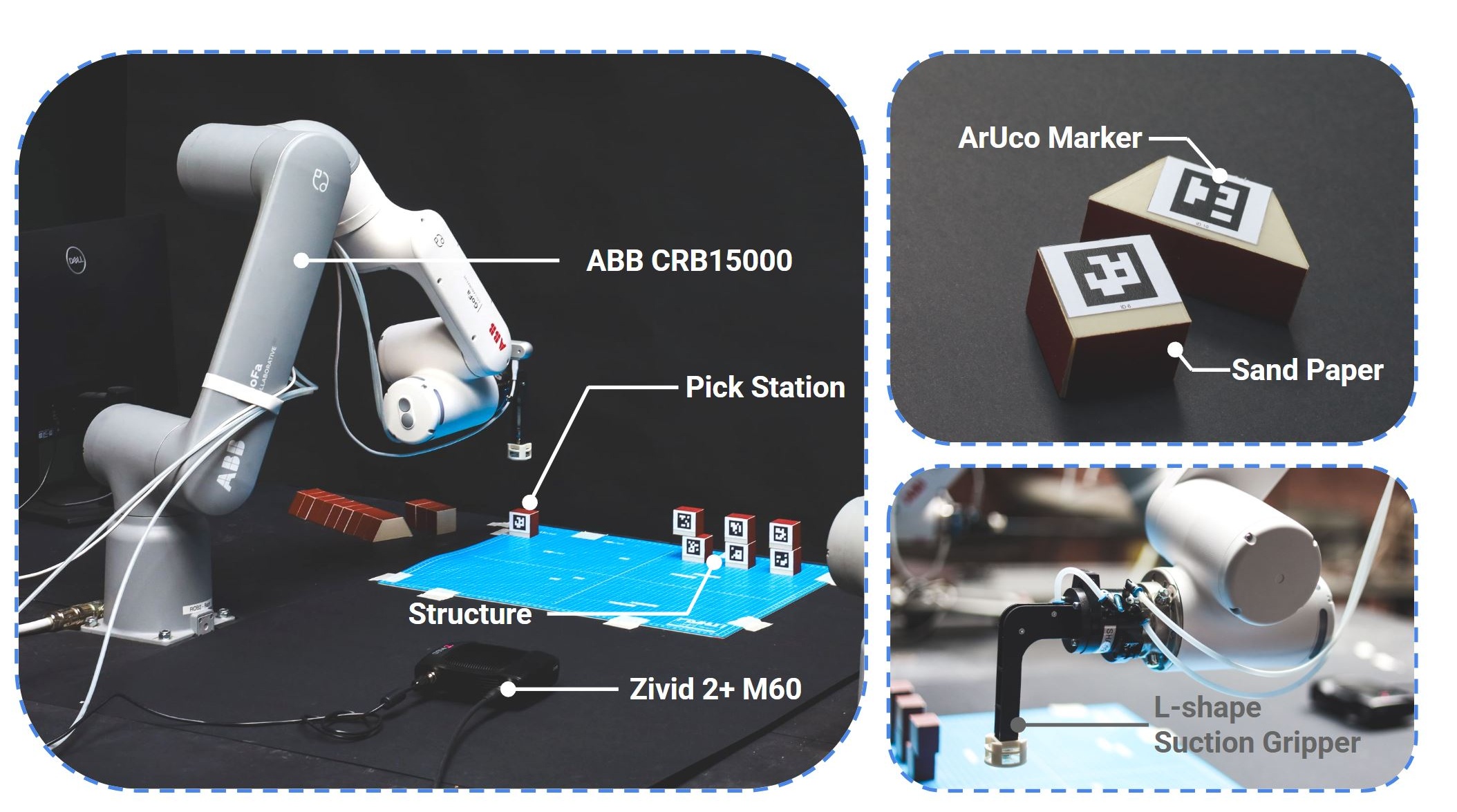}
    \caption{Closed-loop robotic assembly setup: \textbf{Left:} Real-world construction scene with the robotic arm. \textbf{Right Top:} 3D-printed blocks labeled with ArUco markers for visual tracking. \textbf{Right Bottom:} Custom L-shaped suction gripper used for block manipulation.}
    \label{fig:rob_steup}
    \vspace{-1em}
    \end{figure}

As shown in Fig~\ref{fig:rob_steup}, we use an ABB CRB 15000 robotic arm equipped with a suction gripper for handling blocks. %The blocks are 3D printed, and sandpaper is attached to their contact surfaces to increase friction. %to approximate the friction coefficient used in stability simulation.
To monitor construction progress, we establish a visual feedback loop using a Zivid structured-light 3D camera and ArUco markers, as shown in Fig~\ref{fig:workflow}. ArUco markers are chosen as they can be used as 6D pose estimation markers. ArUco markers are printed and affixed to the front face of each block. At each construction step, the Zivid camera scans the assembled structure to obtain point clouds, detects blocks via their ArUco tags, and determines the location of the most recently placed block. This updated state is then passed to the trained policy, which computes the next placement action. The goal of our setup is not only to validate the effectiveness of the trained policy but also the robustness of the construction workflow, and demonstrate its ability to adapt to real-world uncertainties.

\section{Experimental Results}
\label{sec:result}
\subsection{Training and simulation results}

\begin{figure*}[t]
  \vspace{1em}
  \centering
  \includegraphics[width=6in]{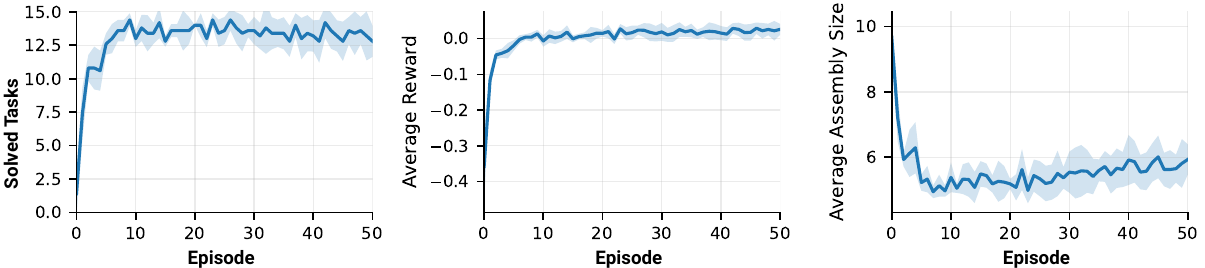}
  \caption{Training performance across five independent random seeds. The plots show the mean number of solved tasks, average cumulative reward, and average assembly size across the 15 training tasks; shaded regions indicate one standard deviation across runs. The policies rapidly improve during the first 10 episodes and subsequently maintain similar performance, solving approximately 13--14 tasks on average. Meanwhile, the average assembly size decreases from about 10 to 5--6 blocks, indicating that the policies learn more material-efficient construction strategies.}
  \label{fig:episode_stats}
  \vspace{-1.5em}
\end{figure*}

We apply our training Algorithm 1 to the 15 tasks in Fig.~\ref{fig:problemdefinition}(right). We train five independent policies using different random seeds, each for 50 episodes, with a discount factor $\gamma=0.9$ and a learning rate of $5\cdot10^{-4}$. Fig~\ref{fig:episode_stats} shows the average number of solved tasks, the cumulative rewards averaged over the 15 tasks, as well as the average number of blocks used per episode throughout the training. As shown in the right plot, the policy gradually learns to construct structures with fewer blocks as training progresses. After 50 episodes, the five training runs achieve similar overall performance, as summarized in Fig.~5. For each training run, we select the first checkpoint that achieves the maximum number of solved tasks reached during that run. One of these checkpoints, obtained at episode 37, solves all 15 training tasks and is used for the subsequent simulation and robotic experiments. The simulated results in Fig.~\ref{fig:success_construct} and Fig.~\ref{fig:failure_construct} show the constructions obtained using this selected policy.

Notably, several tasks are solved using arch-like structures (tasks 8, 9, 12, 14, 15), and task 6 is solved using a counterweight to achieve the necessary overhang to reach the target. This result demonstrates that our algorithm can train a single policy capable of solving a variety of different construction tasks, proposing complex and unintuitive solutions for various types of structures. 

% \begin{figure}
% \centering
% \includegraphics[width=5in]{Figures/final_states.pdf}
% \caption{Constructions generated by the policy in episode 37, which solves all 15 tasks.}
% \label{fig:structure_single}
% \end{figure}

    % \begin{figure}
    % \centering
    % \includegraphics[width=\textwidth]{Figures/trajectory.pdf}
    % \caption{Example of construction trajectory for task 13. The top row shows the assembly and the task data. The second row shows the state image features provided to the model. The bottom row shows the prediction of the successor image, showing the subsequent action sequence the model is planning to take.}
    % \label{fig:trajectory}
    % \end{figure}

%\todo{I feel here might be good to add some positive comments,for example - the above result proves that our algorithm is able to solve multiple defined construction tasks}
% \subsection{Robust construction}

% This section we can show two different things. 
% a) Assume we do not have noise in the environment, when we change the target (obstacle) positions, RL can update its decision based on the new task definition.
% b) Assume we have noise in the environment, we can construct structures with more robustness. 

%\subsection{Transferability}

%Demonstrate that the policy for building a single bridge can be transferred to a double bridge.

\subsection{Baseline comparison}

\begin{table}[t]
\vspace{1em}
\centering
\caption{Node visitation counts for the trained policies and A* across all tasks. 
\textbf{ID}: Task number; 
\textbf{P}: Nodes visited by the trained policy; 
\textbf{A*}: Nodes visited by A* algorithm. 
Task 9 with A* did not succeed, marked by $^\dagger$.
Total nodes visited: P = 5763, A* = 24866 (P/A* = 0.23).
}
\begin{minipage}{0.48\columnwidth}
\centering
% \scriptsize
% \setlength{\tabcolsep}{4pt}
\begin{tabular}{c c c}
\hline
\textbf{ID} & \textbf{P} & \textbf{A*} \\
\hline
1 & 352 & 579 \\
2 & 471 & 1212 \\
3 & 386 & 1725 \\
4 & 604 & 433 \\
5 & 552 & 1143 \\
6 & 317 & 445 \\
7 & 539 & 723 \\
8 & 153 & 999 \\
\hline
\end{tabular}
\end{minipage}
\hfill
\begin{minipage}{0.48\columnwidth}
\centering
% \scriptsize
% \setlength{\tabcolsep}{4pt}
\begin{tabular}{c c c}
\hline
\textbf{ID} & \textbf{P} & \textbf{A*} \\
\hline
9 & 409 & 7299$^\dagger$ \\
10  & 201 & 885 \\
11 & 317 & 999 \\
12 & 309 & 1848 \\
13 & 525 & 4142 \\
14 & 283 & 677 \\
15 & 345 & 1757 \\
\hline
\textbf{$\Sigma$} & \textbf{5763} & \textbf{24866} \\
\hline
\end{tabular}
\end{minipage}
\label{table:node_visited}
\vspace{-1.5em}
\end{table}

We compare our policy against an A* search baseline in the simulation. While both methods rely on the same environment model, A* performs an explicit search over the combinatorial action space, whereas our policy selects among feasible actions using a learned value function. %We measure computational cost by the number of nodes (i.e., states) explored during planning.

A* prioritizes each node using $f(n)=g(n)+h(n)$. The path cost is $g(n)=0.5d(n)$, where $d(n)$ is the number of placed blocks. The heuristic $h(n)$ is the sum of the Euclidean distances from the most recently placed block to all unreached targets.

Each search is limited to 100 expanded nodes and a maximum assembly depth of 10 blocks. During expansion, actions that produce out-of-bounds placements or collisions with existing blocks are rejected, while unstable states are pruned. The search terminates successfully when it finds a stable, collision-free assembly that reaches all targets. It is considered unsuccessful if the frontier is exhausted or the node budget is reached without finding a solution.
We measure computational cost by the number of nodes (i.e., states) explored during planning. 

The result shows that our method achieves a comparable success rate while requiring significantly less computation at test time, requiring approximately 77\% fewer node expansions on average across the 15 tasks (Table. \ref{table:node_visited}).
%In particular, A* requires extensive search to identify feasible action sequences, whereas our policy selects actions in a single forward pass, resulting in substantially faster decision-making. 
Notably, our policy successfully solves all tasks in the simulation, including Task 9, where A* becomes computationally expensive or fails to find a solution within the given budget.
A separate wall-clock time comparison shows a similar overall trend to the node-count results, consistent with RBE stability evaluation being the dominant computational cost during search.

\subsection{Generalization to unseen task configurations}

    \begin{figure}[t]
    \centering
    \includegraphics[width=3in]{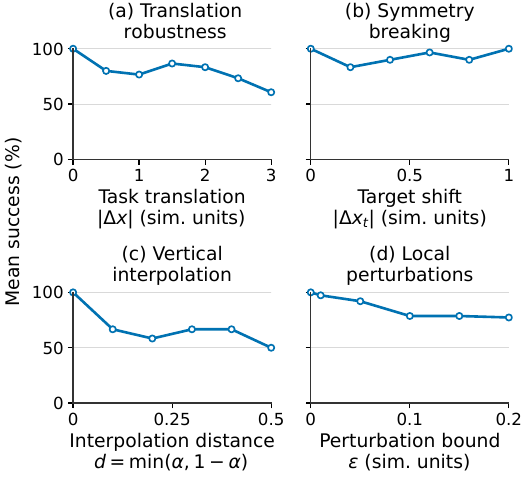}
    \caption{Generalization performance under four types of unseen task variations. (a) Horizontal translation of the entire task, used to evaluate robustness to spatial translation. (b) Symmetry breaking by shifting target locations relative to the obstacles. (c) Interpolation between pairs of training-task geometries, producing unseen intermediate layouts. (d) Local perturbations of target and obstacle positions. The curves report mean task success rate as the magnitude of each variation increases.}
    \label{fig:generalization}
    \vspace{-1em}
    \end{figure}

We evaluate generalization using four types of unseen task variations (Fig.\ref{fig:generalization}). First, to evaluate robustness to spatial translation, we shift the entire task horizontally by translating the targets and obstacles together while preserving their relative geometry. The policy maintains around 80\% success over most translations, with performance decreasing only for large shifts as the task approaches the workspace boundary. Second, we break task symmetry by shifting only the targets relative to the obstacles. This does not consistently increase task difficulty, and success remains above 80\%; in some cases, moving targets away from obstacles enables simpler vertical solutions. Third, we interpolate between pairs of training-task geometries to evaluate unseen intermediate layouts. This is more challenging: success generally decreases as the interpolated task moves away from either source task, reaching approximately 50\% near the midpoint. Finally, under local perturbations of target and obstacle positions, the policy maintains approximately 75--80\% success for moderate perturbations. Overall, the policy generalizes reasonably well to spatial shifts and local geometric variations, but is less reliable for unseen layouts that differ more substantially from the training tasks.

\subsection{Robotic closed-loop construction}

\begin{figure}[t]
    \vspace{1em}
    \centering
    \includegraphics[width=3.3in]{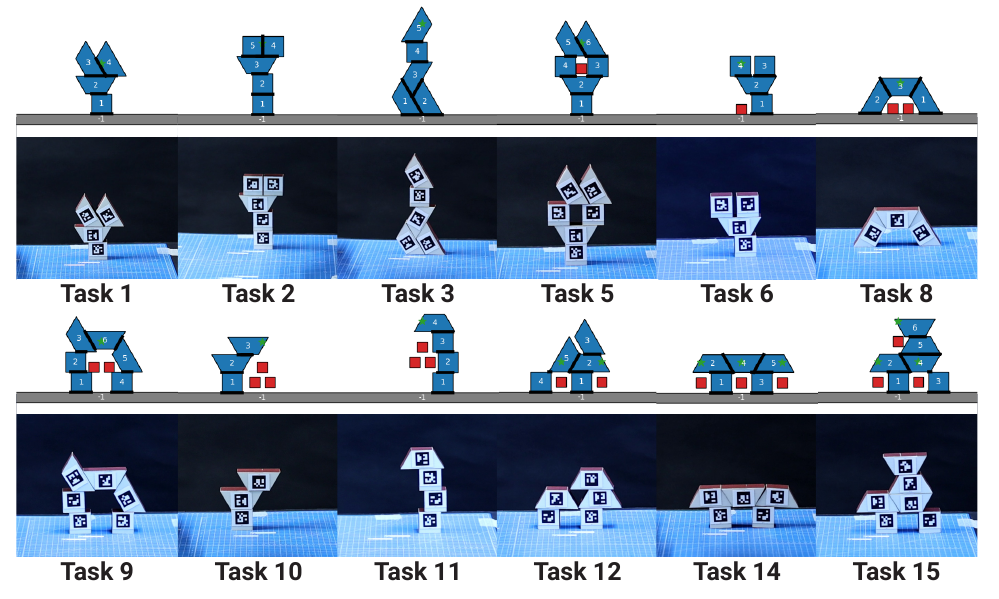}
    \caption{Successfully completed tasks in real-world construction. The top row shows the simulated assemblies, while the bottom row shows the corresponding real-world constructions using closed-loop feedback. Notably, in Task 3 \& 12, the policy produces assemblies that differ from the simulation.}
    \label{fig:success_construct}
    \vspace{-1em}
    \end{figure}

\begin{figure}[t]
    \centering
    \includegraphics[width=3.3in]{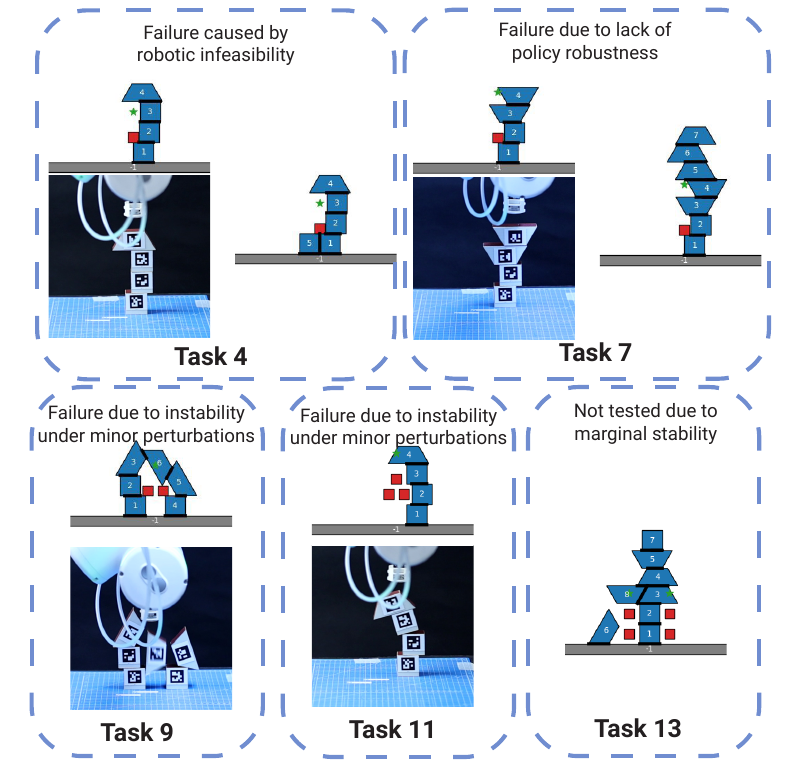}
    \caption{Unsuccessful and untested tasks in real-world construction. Task 4 failed due to robotic infeasibility---specifically, a gripper collision that prevented block placement. Task 7 failed due to limited policy robustness in the presence of accumulated construction noise. Tasks 9 and 11 failed due to structural instability under minor physical perturbations, despite being classified as stable in simulation. Task 13 was not physically tested because its simulated assembly was judged to be marginally stable and difficult to realize reliably in the real-world setup.}
    \label{fig:failure_construct}
    \vspace{-1.5em}
    \end{figure}

We deploy the trained policy on our closed-loop robotic construction system. As shown in Fig~\ref{fig:success_construct}, the robot successfully completes 12 of 15 tasks, 10 of them on the first attempt, as detailed in Table~\ref{table:accuracy}. Two additional tasks (Tasks 9 and 11) are completed after a second trial. Failed tasks are shown in Fig~\ref{fig:failure_construct}. Tasks 4 and 7 do not succeed after multiple attempts, while Task 13 is not physically tested due to the difficulty of achieving its marginally stable assembly.

% Updated construction table

    \begin{table}[t]
    \vspace{0.4em}
    \centering
    \caption{Construction results for all tasks. 
    \textbf{ID}: Task number; 
    \textbf{S}: Successful trials / Total trials; 
    \textbf{Avg}: Average relative placement offset; 
    \textbf{Max}: Maximum relative placement offset. Offsets are normalized by block size. Note that we did not test Task 13.}
    \begin{minipage}{0.48\columnwidth}
    \centering
    \begin{tabular}{c c c c}
    \hline
    \textbf{ID} & \textbf{S} & \textbf{Avg} & \textbf{Max} \\
    \hline
    1  & 1/1 & 0.04 & 0.08 \\
    2  & 1/1 & 0.01 & 0.03 \\
    3  & 1/1 & 0.03 & 0.08 \\
    4  & 0/4 & 0.03 & 0.06 \\
    5  & 1/1 & 0.03 & 0.08 \\
    6  & 1/1 & 0.01 & 0.03 \\
    7  & 0/6 & 0.04 & 0.09 \\
    
    \hline
    \end{tabular}
    \end{minipage}
    % \hspace{1cm}
    \begin{minipage}{0.48\columnwidth}
    \centering
    \begin{tabular}{c c c c}
    \hline
    \textbf{ID} & \textbf{S} & \textbf{Avg} & \textbf{Max} \\
    \hline
    8  & 1/1 & 0.01 & 0.01 \\
    9   & 1/2 & 0.02 & 0.05 \\
    10  & 1/1 & 0.01 & 0.01 \\
    11  & 1/2 & 0.03 & 0.05 \\
    12  & 1/1 & 0.01 & 0.04 \\
    14  & 1/1 & 0.01 & 0.03 \\
    15  & 1/1 & 0.03 & 0.04 \\
    \hline
    \end{tabular}
    \end{minipage}
    \label{table:accuracy}
    \vspace{-1.5em}
    \end{table}

% Notably, in two successful cases, Task 3 and Task 12, the constructed assemblies are different from the simulated results. These variations arise from real-world construction noise and illustrate the policy’s capacity to adapt to unforeseen deviations, even though it is trained without explicit noise modeling. This adaptability highlights the generalizability of the learned policy.

Failures in robotic construction arise from three main factors. First, in one long-horizon task (Task 7), small placement errors accumulate across many steps and result in policy failure. While the closed-loop system can correct moderate deviations, its adaptability has limits; beyond a certain threshold, the drift becomes too large for the policy to recover. 
%First, there is one case (Task 7) that with relative long assembly sequences, small errors accumulate due to construction noise, leading to policy failure. 
Second, our binary stability solver (RBE) cannot capture marginally stable cases that collapse under slight disturbances, as in Tasks 9 and 11. In Task 13, for instance, Block 8 is theoretically supported by compressive forces from Block 3 and 4 but can easily fail under perturbation in reality. Third, hardware constraints limit feasibility. In Task 4, the gripper’s geometry prevents placing Block 5 without collision. This issue highlights the need to integrate reachability and collision checks into policy learning, and an accurate and robust numerical simulation of assembly stability.

Notably, in two successful cases, Task 3 and Task 12, the constructed assemblies differ from the simulated outcomes. These deviations result from real-world construction noise and demonstrate that the policy can adapt its strategy when the physical structure evolves differently than expected. This capability is especially important in construction, where material tolerances, sensing errors, and on-site uncertainties are unavoidable. This kind of adaptive behavior is precisely what is needed to deploy autonomous methods in practical construction environments.

To quantify construction noise, we compare the policy-generated actions with the actual placements executed by the robot, and report the average and maximum positional offsets normalized by the block size in Table~\ref{table:accuracy}. Several factors contribute to the construction noise: unevenness of the construction table, imperfect tool center point calibration, slight misalignment when attaching ArUco markers, and minor deviations in block placement at the pick station. Despite these sources of noise, the overall deviation remains relatively small. As a result, the trained policy, even without explicit modeling of construction noise, can be reliably deployed on a real robot to accomplish 80\% of tasks.

% Offsets are expressed as a ratio relative to the square block edge length, providing a scale-invariant measure of construction accuracy. Using relative values rather than absolute values allows for a clearer interpretation of precision in relation to the overall size of the structure.

\section{Conclusion}
\label{sec:conclusion}
In summary, our work makes the following contributions:

\begin{itemize}
    \item We propose a novel robotic construction framework that enables the autonomous assembly of simple 2D stable structures without relying on predefined architectural plans.

    \item We formalize construction tasks in terms of geometric goals, defined by targets and obstacles, rather than fixed structural forms. This allows layouts to emerge during the building process.
    
    \item We develop a goal-conditioned RL approach that uses deep Q-learning with successor features to solve multiple construction tasks using a single policy. We rely on an image-based successor-feature formulation that separates anticipated future construction from the task-dependent reward, providing a spatially interpretable representation of the policy's long-term construction behavior, accommodates polygon-shaped blocks in a shared representation, and supports translation robustness within the workspace, although full translational equivariance is broken by the workspace boundaries. This representation enhances the interpretability of the learned policy by making its anticipated long-term construction behavior visually traceable.

    \item We validate the framework on 15 construction tasks, both in simulation and on a real-world closed-loop robotic assembly process. Across five independent training runs, the policies solve approximately 13--14 of the 15 training tasks on average, while the selected checkpoint solves all 15 tasks. In physical experiments, the selected policy successfully completes 12 of the 14 attempted tasks despite construction noise. These results highlight the potential of our framework for more flexible and adaptive robotic construction in dynamic, real-world environments.

\end{itemize}

\subsection{Limitations and future work}
% \textbf{Design Intention Alignment.}
% By removing predefined blueprints, the system is free to discover its own construction strategies, which may differ from what a human designer would anticipate. In Task 11, for instance, the policy generated a structurally valid solution that did not match our initial design intention. Rather than a drawback, this highlights the generative capacity of the RL approach. Moving forward, a more interactive setup could allow human designers to inject preferences or constraints, enabling collaborative decision-making while preserving the flexibility of the learned policy.

% \textbf{Limited Design Space.}  
% Currently, our work operates in 2D with only two types of blocks and one robotic arm. This simplification limits the action space of the reinforcement learning agent, allowing for feasible training times. As a proof of concept, this work demonstrates the feasibility of blueprint-free construction; future work will extend the approach to more diverse block shapes, full 3D construction tasks, and with multiple robotic arms, which should unlock more complex assemblies.

\textbf{Limited Design Space.} Currently, our work operates in 2D with two block types and one robotic arm. This simplification keeps the action space and training cost tractable, but limits the complexity of the assemblies considered. Extending the framework to larger structures, additional block types and orientations, or full 3D construction would substantially increase the number of candidate placements and the cost of stability evaluation. In our current implementation, RBE evaluation is already the main computational bottleneck during training. As a proof of concept, this work demonstrates the feasibility of blueprint-free construction; future work will therefore focus not only on richer 3D assembly settings, but also on more efficient and parallel stability evaluation, potentially through GPU-based implementations, to make such scaling computationally practical.

\textbf{Sim-to-Real Gap.}
As demonstrated in our closed-loop robotic constructions, a noticeable sim-to-real gap remains, arising from three main factors. First, the stability solver (RBE) provides only binary outputs, so marginally stable structures may fail under real-world disturbances. Future work will explore more accurate solvers~\cite{kao_coupled_2022}. 
Second, actions produced by the RL policy may be infeasible because of robotic constraints such as gripper collisions and limited reachability; incorporating these constraints during training should improve deployment feasibility. Third, the current policy does not account for construction noise or uncertainty. Although it achieves high success rates, performance declines in long-horizon tasks. Our RL framework enables the inclusion of real-world noise during training, which we plan to explore to develop more robust and reliable policies.

\textbf{Long-Term Vision.}
Looking ahead, we envision this framework as a foundation for autonomous construction systems capable of working with diverse materials and scales. Expanding beyond current 3D-printed blocks, future developments could incorporate natural, irregular, or recycled materials, a step toward real-world architectural applications. The ability to assemble structures without predefined plans also opens opportunities for remote or difficult-to-access scenarios, such as rapid post-disaster reconstruction. Ultimately, our goal is to move toward autonomous systems that can support architectural-scale fabrication while remaining adaptable to the uncertainties inherent in construction sites.

\bibliographystyle{IEEEtran}
\bibliography{mybibfile}

\newpage

\vfill

\end{document}